\documentclass[letterpaper, 10pt, conference]{IEEEtran}
\usepackage[letterpaper, top=54pt, bottom=54pt, left=54pt, right=54pt]{geometry}

\IEEEoverridecommandlockouts

\usepackage{cite}
\usepackage{amsmath,amssymb,amsfonts}
\usepackage{graphicx}
\usepackage{textcomp}
\usepackage{xcolor}
\usepackage{booktabs}
\usepackage{multirow}
\usepackage[hidelinks]{hyperref}

\usepackage{xcolor}
\usepackage{balance}

\begin{document}

\title{\LARGE \bf Autonomous Telerehabilitation via Skeletal Motion\\Prediction and Joint-Level Performance Assessment\\}

\author{
  \IEEEauthorblockN{Lara Pereira, João Ruivo Paulo, Pedro Santos and Paulo Peixoto}
  \IEEEauthorblockA{
    \textit{Institute of Systems and Robotics}\\
    \textit{Department of Electrical and Computer Engineering, University of Coimbra, Portugal}\\
    {\{lara.pereira,peixoto,jpaulo\}@isr.uc.pt}
  }
\thanks{© 2026 IEEE. Personal use of this material is permitted. Permission from IEEE must be obtained for all other uses, in any current or future media, including reprinting/republishing this material for advertising or promotional purposes, creating new collective works, for resale or redistribution to servers or lists, or reuse of any copyrighted component of this work in other works.}  
}

\maketitle


\begin{abstract}
Autonomous rehabilitation systems must not only recognize human motion but also provide structured feedback to support users without continuous therapist supervision. This paper presents a telerehabilitation pipeline that integrates skeleton-based exercise quality assessment and
short-term motion prediction into a two-module system operating on marker-free RGB video. A self-attentive Bidirectional LSTM performs exercise quality classification using MMD-NCA metric learning, while a graph-based motion prediction module computes per-joint position errors between predicted and observed poses, generating spatially localized deviation signals. Each module is evaluated independently on established benchmarks: the classifier achieves 96.45\% mean-class accuracy on squat
sequences from the PROZIS dataset, and the adopted STARS predictor achieves a mean MPJPE of 75.8\,mm at 560\,ms on Human3.6M, outperforming graph and recurrent baselines across all prediction horizons. The framework is designed for eventual deployment in assistive robotics and home-based
rehabilitation contexts; end-to-end integration and clinical validation are important directions for future work. By combining motion recognition and prediction in a single system, this work contributes a step toward autonomous, feedback-driven telerehabilitation, for more accessible and scalable rehabilitation solutions.

\end{abstract}


\section{Introduction}

In the past decade, healthcare has undergone a significant transformation driven by digital technologies, particularly in the domain of physical rehabilitation~\cite{Rigamonti2020}. With the rising prevalence of musculoskeletal conditions and the growing demand for long-term care, telerehabilitation has emerged as a scalable solution that enables patients to perform therapy remotely, reducing travel costs and increasing accessibility ~\cite{Liao2020}.

However, current telerehabilitation systems largely rely on passive video guidance and lack the ability to provide structured, personalized feedback. This limitation restricts their effectiveness, as patients may
perform exercises incorrectly without correction, potentially reducing therapeutic outcomes. In contrast, human therapists continuously observe, evaluate, and guide patient movements, highlighting the need for automated systems that can detect movement deviations and communicate them to the user, even though fully replicating the therapist’s clinical reasoning remains a long-term goal.

In the context of human–robot interaction and assistive robotics, this challenge translates into enabling intelligent systems, such as companion or social robots, to perceive, interpret, and respond to human motion in a meaningful and user-centered manner. For such systems to support rehabilitation effectively, they must not only assess whether an exercise is performed correctly but also anticipate how the movement should evolve over time. This predictive capability allows the system to compare expected and observed motion, enabling spatially localized, joint-level feedback that can be directly communicated to the user.

To address this, we propose a machine learning-based pipeline for telerehabilitation support that operates on marker-free RGB video. The framework integrates exercise quality assessment and short-term motion prediction, enabling the system to generate both holistic
(correct\,/\,incorrect) labels and spatially localized deviation signals at the joint level. Each module is validated on established benchmarks, and their combined design is intended to support future deployment in human-centered robotic assistance scenarios.

 The contributions of this work are:
\begin{itemize}
  
  \item An integrated two-module pipeline for telerehabilitation support that combines skeleton-based exercise quality assessment and short-term motion prediction from marker-free RGB video. The primary contribution is the integration of holistic quality classification and joint-level deviation signals into a single system design, intended as a component of a future human--robot rehabilitation system.
  \item Evaluation of MMD-NCA distributional metric learning on an annotated rehabilitation exercise dataset, achieving 96.45\% mean-class accuracy on squat sequences.
  \item A joint-level position error (JPE) formulation and color-coded visualization scheme for spatially localized rehabilitation feedback.
  \item Characterization of data requirements for generalizing skeleton-based quality classifiers across exercise types.
\end{itemize}

We position this work as establishing the technical feasibility of the individual perception and feedback-generation components required for an autonomous rehabilitation coach, rather than as a complete, evaluated human-robot interaction system. Integration with a robot or interactive front-end, and evaluation with users and therapists, are identified as the necessary next steps toward the human-centered scenario motivating this work.

\begin{figure*}[!t]
  \centering
  \includegraphics[width=\textwidth,trim=0cm 0cm 0cm 2.8cm, clip]{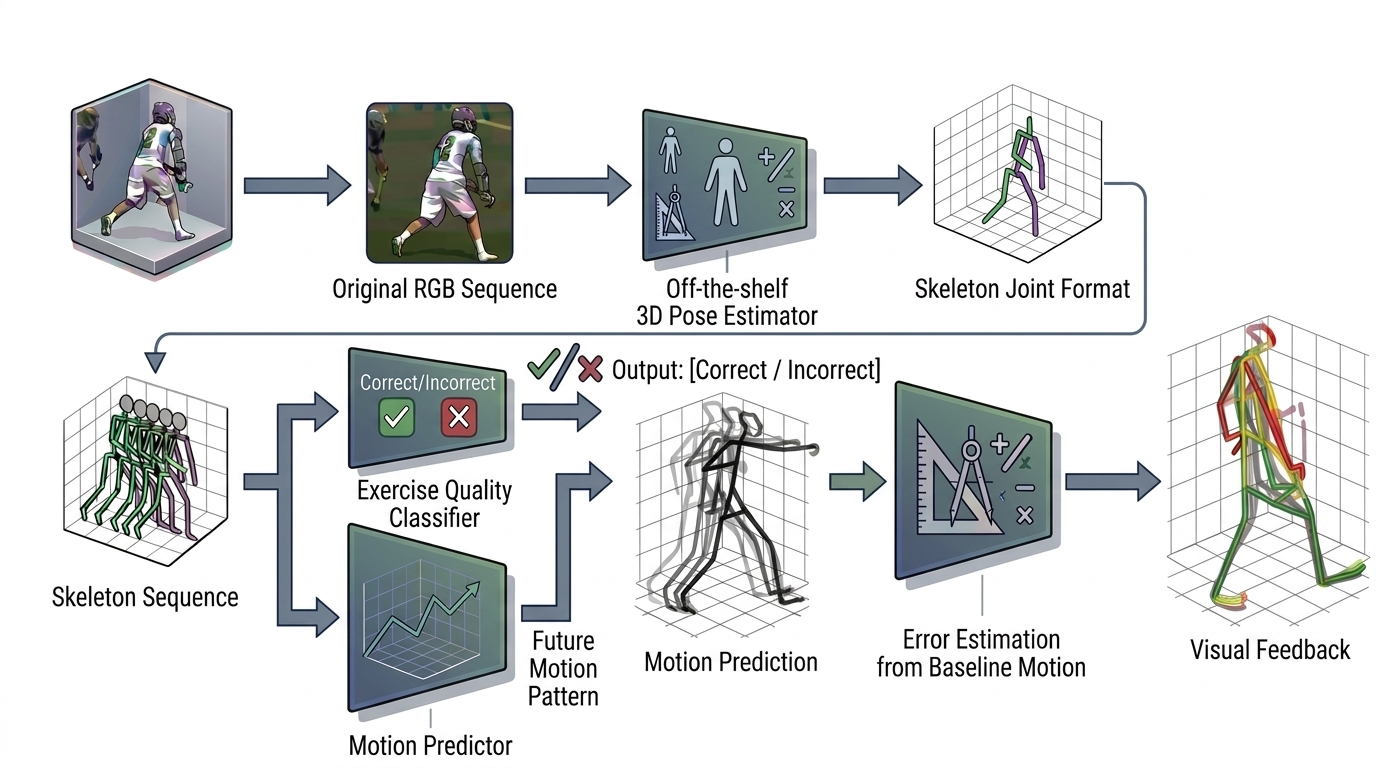}
  \caption{High-level pipeline of the proposed method.}
  \label{fig:HLP}
\end{figure*}
\section{Related Work}

Automated assessment of rehabilitation exercise quality from skeleton or video data has been studied using both rule-based and learning-based approaches. Deep learning methods have largely replaced traditional signal processing techniques because of their ability for end-to-end feature learning and adaptability to inter-subject variability. A key challenge is that human motion sequences of the same exercise type can vary substantially in length and tempo, requiring architectures or preprocessing strategies that are robust to these temporal variations.

Bidirectional LSTMs have shown strong performance on sequential pose data by capturing both forward and backward temporal context. A persistent challenge in motion sequence comparison is that direct distance measures do not account for temporal misalignment, while Dynamic Time Warping (DTW) \cite{vintsyuk1968speech, Keogh2001} addresses alignment but struggles with subtle temporal variations and scales poorly to high-dimensional data \cite{Coskun2018n, Keogh2001}.

Metric learning approaches address this by learning an embedding in which motion sequences from the same class cluster together. Hadsell \textit{et al.}~\cite{Hadsell2006n} used Siamese networks with contrastive loss, but they only exploit pairwise relationships. Triplet loss \cite{Schroff2015n} improved upon this by considering anchor, positive, and negative samples jointly, but it requires careful negative mining and sensitive margin hyperparameter tuning. Neighborhood Components Analysis (NCA) and its distributional extension, Maximum Mean Discrepancy NCA (MMD-NCA) \cite{Coskun2018n}, eliminate these fixed margins by normalizing over negative classes in the batch. This progression provides a robust foundation for handling variable-length action recognition without the drawbacks of explicit time-warping algorithms.

In motion prediction, with a continued focus on skeleton data, early Recurrent Neural Network (RNN)-based approaches to human motion prediction established key architectural patterns. For example, Martinez \textit{et al.}~\cite{Martinez2017n} framed short-term prediction as a sequence-to-sequence problem, using separate encoders and decoders to improve alignment between training and inference protocols. 
However, Graph Convolutional Networks (GCNs) were subsequently proposed for skeleton-based understanding by Yan \textit{et al.}~\cite{Yan2018n}, proving highly effective at encoding joint connectivity via learnable adjacency matrices.
Building on this, Mao \textit{et al.}~\cite{Mao2020} established a strong baseline using Discrete Cosine Transform (DCT)-domain graph networks (LTD) with explicit trajectory dependencies. Recent advancements have focused on increasing the efficiency and diversity of these predictions. Architectures like Space-Time-Separable GCN (STS-GCN)~\cite{Sofianos2021n} introduced factorized spatial and temporal processing to reduce computational overhead, while methods such as Spatial-Temporal Anchor-Based Sampling (STARS)~\cite{Xu2022n} incorporated generative modeling to capture the inherent multimodal nature of future human motion.

Despite rapid advancements in individual tasks, few works integrate both action recognition and motion prediction into a single pipeline for rehabilitation feedback. This gap motivates the present work.

\section{Pipeline}
\label{sec:pipeline}

The proposed approach consists of three main modules, as shown in Fig.~\ref{fig:HLP}. An off-the-shelf human pose estimator (\textit{MediaPipe} or \textit{OpenPose}) converts an input RGB sequence into a time series of 3D skeleton joint coordinates. The skeleton is normalized relative to the hip joint at each frame, removing absolute position and making the representation invariant to subject placement within the camera
frame. Variable-length repetition sequences are either subsampled to a maximum length or zero-padded to a minimum length before being passed to the two processing modules. Interpolation was also evaluated but yielded marginally inferior results ($\leq$2 percentage points in mean-class accuracy) compared to zero-padding.

The normalized skeleton sequence is processed by two parallel modules. The \emph{exercise quality classifier} (Section~\ref{sec:ar}) operates on
full-repetition sequences and produces a binary holistic quality label (correct or incorrect) for each repetition. The \emph{motion predictor} (Section~\ref{sec:mp}) uses a short context window of 10 frames (400\,ms) and predicts the next 5--25 frames (200--1000\,ms). The predicted joint positions are compared to the observed ground truth using a per-joint Euclidean error, generating a spatially localized error map. This map is thresholded into four color categories: green (low error), yellow (moderate), orange (elevated), and red (high deviation), and rendered as an overlay on a 3D skeleton visualization. Joints irrelevant to the exercise under assessment (e.g., arms during a gait task) are masked to focus feedback on clinically relevant body parts.

A key design requirement of the proposed system is robustness to variations in exercise tempo, which are common in real-world rehabilitation scenarios, especially among older adults and users with limited mobility. To ensure consistent performance across different movement speeds, the BiLSTM module uses a self-attention mechanism that aggregates variable-length sequences into a fixed-length representation, enabling tempo-invariant assessment. In parallel, the graph-based prediction models use trainable temporal adjacency matrices to implicitly learn speed-adaptive motion patterns directly from data, eliminating the need for explicit temporal normalization or dynamic time warping. 
Together, these design choices support reliable performance and feedback across diverse user behaviors and execution styles.

\section{Exercise Quality Classification}
\label{sec:ar}
Exercise quality takes the 3D skeletons as input with the goal of classifying an exercise as correctly or incorrectly executed. Below we describe the adopted approach for this action recognition task.

\subsection{Layer-Normalised Bidirectional LSTM}
Here, we adopted the Coskun \textit{et al.}'s~\cite{Coskun2018n} approach using a BiLSTM with a self attention mechanism was chosen.
The core building block is a layer-normalized LSTM (LNLSTM) that performs forward and backward passes over the input pose sequence $X = (x_1, \ldots, x_n)$, where $x_t \in \mathbb{R}^{J \times 3}$ is the pose at timestep $t$. Layer normalization replaces batch normalization inside the LSTM cell,
which is known to degrade performance in standard LSTM implementations \cite{Coskun2018n}. The normalized hidden state is computed as:
\begin{equation}
  g^m_t = \frac{1}{H}\sum_{j}^{H} g^{c,j}_t, \quad
  v_t = \sqrt{\frac{1}{H}\sum_{j}^{H}(g^{c,j}_t - g^m_t)^2}
  \label{eq:layernorm}
\end{equation}
\begin{equation}
  h_t = g^o_t \!\odot\! \tanh\!\Bigl(
    \tfrac{\gamma}{{v_t}} \!\odot\! (g^c_t - g^m_t) + \beta
  \Bigr)
  \label{eq:lnlstm}
\end{equation}
where $g^c_t$ and $g^o_t$ are the cell memory and output gate, $H{=}128$
is the number of hidden units, and $\gamma, \beta \in \mathbb{R}^H$
are learnable scale and bias parameters. The forward and backward LNLSTM
outputs are concatenated at each timestep as $s_t = [s_{t,f};\, s_{t,b}]$,
producing the sequence of states $S = \{s_1,\ldots,s_n\}$.

\subsection{Self-Attention Mechanism}

In the context of human motion sequences, certain poses intuitively convey more information than others.
To prioritize the most relevant postures, we use a self-attention mechanism that, based on the work of Lin~\textit{et al.}~\cite{Lin2017}, assigns a scalar importance weight to each timestep:
\begin{equation}
  r = W_{s_2}\tanh(W_{s_1}S^T), \quad
  a_i = -\log\!\Bigl(\tfrac{e^{r_i}}{\sum_j e^{r_{i,j}}}\Bigr)
  \label{eq:attn}
\end{equation}
with weight matrices $W_{s_1} \in \mathbb{R}^{200 \times 10}$ and
$W_{s_2} \in \mathbb{R}^{10 \times 1}$, 
where \textit{r} has values in the range of [0,1] and is used to give different weights to $a_i$. 

The attended, fixed-length
sequence embedding $S_C = r \cdot a \in \mathbb{R}^{128}$ is then passed
to a classification head consisting of:
FC(320) $\!\to\!$ Dropout(0.5) $\!\to\!$ BatchNorm $\!\to\!$
FC(320) $\!\to\!$ BatchNorm $\!\to\!$ FC(128) $\!\to\!$ BatchNorm $\!\to\!$ $L_2$-Norm,
with ReLU activations on all but the final fully connected (FC) layer.

All square weight matrices are initialized with random orthogonal matrices;
remaining matrices use a uniform distribution ($\mu{=}0$, $\sigma{=}0.001$).

\subsection{MMD-NCA Loss Function}
The architecture uses a variation of the NCA, based on the Maximum Mean Discrepancy (MMD). The MMD-NCA loss \cite{Coskun2018n} operates at the distributional level, measuring divergence between class embedding distributions via the MMD. Given sample sets $X {=} \{x_1,\ldots,x_m\}$ and
$Y {=} \{y_1,\ldots,y_n\}$ and a mixture of $K$ Gaussian kernels $k$:
\begin{equation}
\begin{split}
    \textrm{MMD}[k,X,Y]^2 &= \frac{1}{m^2}\sum_{i=1}^{m}\sum_{j=1}^{m}k(x_i,x'_j) \\
    &\quad - \frac{2}{mn}\sum_{i=1}^{m}\sum_{j=1}^{n}k(x_i,y_j) \\
    &\quad + \frac{1}{n^2}\sum_{i=1}^{n}\sum_{j=1}^{n}k(y_i,y'_j)
\end{split}
\label{eq:54}
\end{equation}
To differentiate between correct and incorrect exercise executions, the network uses the MMD-NCA loss proposed by Coskun \textit{et al.}~\cite{Coskun2018n}. This function effectively reduces the overlap between category distributions in the embedding space while ensuring that samples from the same distribution remain as close as possible. Specifically, the final loss encourages the anchor embedding distribution to match the positive class distribution, while being separated from $M$ randomly sampled negative classes:

\begin{equation}
\mathcal{L}_{\text{MMD-NCA}} =
\frac{\exp(-\text{MMD}[k, f(X), f(X^+)])}
{\sum_{j=1}^{M}\exp(-\text{MMD}[k, f(X), f(X_{cj}^{-})])}
\label{eq:mmdnca}
\end{equation}
where $X$ and $X^+$ denote motion samples belonging to the same action category, respectively. The term $X_{cj}^{-}$ represents samples from one of $M$ distinct, randomly selected categories $c_j \in C$, and $f(\cdot)$ denotes the network's mapping to the low-dimensional embedding space.

\subsection{Training}
The model is trained with a batch size of 32 using the Adagrad optimizer and an initial learning rate of $10^{-3}$. A learning rate scheduler is applied, reducing the rate by a factor of 0.8 with a patience of 7 epochs across approximately 1000 training sequences. Binary cross-entropy serves as the classification objective alongside the MMD-NCA embedding loss. Training ran for up to 1000 epochs (with convergence typically observed before the 500th epoch) on an NVIDIA RTX 2060 GPU ($\approx$4.2\,s per epoch).

\section{Short-Term Motion Prediction}
\label{sec:mp}
To further quantify the evaluation of human movement in a rehabilitation setting, another framework was proposed: a network that could estimate or predict the next \textit{x} frames based on an average model, then compare them to the ground truth, allowing feedback at the joint level rather than just through an overall scoring function. For this purpose, two graph-based networks were chosen.

\subsection{Problem Formulation}

Let $\mathbf{X}_\text{in} = [X_1,\ldots,X_T]$ denote the observed pose history, where $X_i \in \mathbb{R}^{3 \times V}$ encodes the 3D coordinates
of $V$ joints for the time sequence \(i=1,...,T\). 
The goal is to predict $\mathbf{X}_\text{out} = [X_{T+1},\ldots,X_{T+K}]$, the poses for the next $K$ frames. Both models use $T{=}10$ context frames (400\,ms) and are evaluated across two distinct prediction horizons: 
\textbf{short-term} predictions from $K{=}2$ to $K{=}10$ (80--400\,ms), and \textbf{long-term} predictions from $K{=}14$ to $K{=}25$ (560--1000\,ms).

\subsection{Space-Time-Separable GCN (STS-GCN)}

STS-GCN \cite{Sofianos2021n} encodes the full observation as a spatio-temporal graph \(\mathcal{G}=(\mathcal{V},\mathcal{E})\) with $TV$ nodes. The key insight is to factorize the full spatio-temporal adjacency matrix $A^{st} \in \mathbb{R}^{VT \times TV}$
into separate spatial and temporal components:
\begin{equation}
  \mathcal{H}^{l+1} = \sigma(A^{s-l} A^{t-l} \mathcal{H}^l W^l)
  \label{eq:stsgcn}
\end{equation}
where $A_s \in \mathbb{R}^{V \times V}$ captures joint-joint relations,
$A_t \in \mathbb{R}^{T \times T}$ captures time-time relations, and
$W^l$ are the trainable convolutional weights at layer $l$. This factorization limits joint-time cross-talk, reducing the number of
learnable parameters while preserving the expressiveness of the full
space-time representation. Four encoder layers with batch normalization and residual connections are used. A Temporal Convolutional Network (TCN) decodes the latent representation into predicted future joint coordinates.
Training minimized Mean Per Joint Position Error (MPJPE) loss using Adam (learning rate 0.001, decay factor 0.1 with patience 5 epochs), batch size 256, for 30~epochs on an RTX~2060 (20--30 minutes total).

\subsection{Spatial-Temporal Anchor-Based Sampling (STARS)}

Following Xu \textit{et al.}~\cite{Xu2022n}, we adopt the STARS framework, which extends the STS-GCN backbone with a generative model that decomposes the latent code into a stochastic component $z \sim p(z)$ and a set of $K$ learnable deterministic anchors $a = \{a_k\}_{k=1}^K$. Anchors are further factorized into $K_s$ spatial anchors $\{a^s_i\}$ (controlling movement direction and posture) and $K_t$ temporal anchors $\{a^t_j\}$ (controlling frequency and speed), yielding $K_s \times K_t$ compositional predictions:
\begin{equation}
  \hat{Y}_k = \mathcal{G}(a^s_i + a^t_j,\, z,\, X)
  \label{eq:stars}
\end{equation}
The backbone operates in the DCT frequency domain using 8 STS-GCN layers (four standard and four pruned) with alternating channel dimensions (3$\to$128$\to$64$\to$128$\to$64$\to$128$\to$64$\to$128$\to$3). 
It utilizes cross-layer adjacency sharing and kinematic-tree-guided spatial pruning of the spatial adjacency matrix. 

Three loss terms are combined during training: a reconstruction loss (minimizing the best prediction against ground truth), a diversity-enhancing loss (preventing anchor collapse),
and motion constraint losses (history reconstruction error, pose prior,
limb loss, and angle loss). 

The learning rate is set to 0.001 and decays linearly after epoch 100. In this work, we use $K{=}1$ to produce a single predicted trajectory per context window.

\subsection{Joint-Level Position Error}
To generate spatially and temporally localized feedback, we compute a per-joint, per-frame position error (JPE). By omitting the averaging over both joints and time present in standard MPJPE, we isolate the specific error for each articulation at every individual time-step:
\begin{equation}
  \mathcal{L}_\text{JPE}(v, k) = \left\|\hat{x}_{vk} - x_{vk}\right\|_2
  \label{eq:jpe}
\end{equation}
where $\hat{x}_{vk}$ and $x_{vk}$ are the predicted and observed 3D positions of joint $v$ at timestep $k$. This yields a scalar error for each joint in each predicted frame, which is then mapped to four color thresholds for visualization. 
Joints not relevant to the exercise being assessed are masked in the final overlay.

\section{Datasets}
\label{sec:data}

To evaluate the proposed telerehabilitation framework, we consider multiple datasets that capture complementary aspects of the problem, including controlled motion prediction benchmarks and real-world rehabilitation exercise data. This combination allows us to assess both the technical performance of the underlying models and their applicability to practical, user-centered rehabilitation scenarios.

Specifically, we employ the Human3.6M dataset to benchmark motion prediction under standardized conditions, enabling comparison with prior work, and the CMU Motion Capture dataset to evaluate the robustness of the action recognition and metric learning components across diverse motion categories. In addition, we use the PROZIS Challenge dataset, which contains annotated fitness exercises performed by users with varying levels of expertise, to assess performance in a realistic telerehabilitation context.

Together, these datasets provide a comprehensive evaluation framework, spanning controlled laboratory settings and real-world exercise variability, and supporting the validation of both system-level performance and its relevance to assistive rehabilitation applications.
 
\subsection{Human3.6M}
Human3.6M \cite{Ionescu2014n} is a large-scale 3D human motion capture dataset recorded at 50\,Hz with a Vicon system, comprising 7 subjects performing 15 everyday actions. We use a 22-joint subset consistent with prior work~\cite{Sofianos2021n, Mao2020}. Two parameterizations are evaluated: 3D Cartesian coordinates (requiring no additional preprocessing) and 3D Euler angles converted to exponential map format using the preprocessing pipeline of Jain \textit{et al.}~\cite{Jain2016n}, which avoids gimbal locking artifacts. Five subjects are used for training, one for validation, and one for testing.

\subsection{PROZIS Challenge Dataset}
 
The PROZIS Challenge dataset~\cite{batista2019prozis, Ferreira2022} is a proprietary collection of annotated fitness exercise videos gathered in 2019.

It covers five exercise types: squats, sit-ups, push-ups, jumping jacks, and burpees, performed by participants with varying fitness levels. Each repetition is labeled as correctly or incorrectly executed. 3D skeletal landmarks were extracted from video and normalized relative to the hip joint. Repetition boundaries were manually annotated by segmenting each sequence into up and down phases.

The squat class has the largest sample count (approximately 950 sequences), while other exercise types range from about 70 to 300 sequences.
Sequences of variable length are zero-padded or subsampled to a fixed target length and split 70/15/15 into training, validation, and test sets. An important limitation is that the dataset lacks joint index metadata and explicit kinematic tree information, which prevents the use of graph-based architectures and restricts this dataset to the sequence-level BiLSTM classifier.
This constraint is also why the classification and motion-prediction modules are evaluated on separate benchmarks in Section~\ref{sec:results} rather than jointly on a single dataset: PROZIS cannot currently support the graph-based predictor, and Human3.6M contains no annotated rehabilitation exercises. Closing this gap is identified as a primary direction for future work (Section~\ref{sec:conclusion}).

\subsection{CMU Dataset}
The CMU Graphics Lab Motion Capture Database \cite{CMUd} is used to benchmark the action recognition and metric learning baselines. The raw sequence data is preprocessed by downsampling the original 120\,Hz capture rate to 30\,Hz and removing six redundant joints. To prevent gimbal locking artifacts during training, the remaining 3D Euler angles are converted to an exponential map representation \cite{Coskun2018n, Sutherland2021n}. Of the 38 available action categories, a balanced split is used, with 19 categories for training and the remaining 19 distinct categories for testing.

\section{Results}
\label{sec:results}

This section evaluates the proposed telerehabilitation framework in terms of technical performance and suitability for real-world, user-centered rehabilitation scenarios. Given the system's dual nature, we assess the exercise quality classification and motion prediction modules independently, as each was trained and evaluated on separate benchmarks optimized for its respective task. For the classification component, we focus on metrics that address class imbalance and decision-making reliability, while for motion prediction, we evaluate spatial accuracy across varying temporal horizons.

\subsection{Evaluation Metrics}

Since the action recognition task is a classification task and the PROZIS dataset is unbalanced, it is important to select metrics that accurately represent the overall performance of the method. Therefore, we report mean-class accuracy (mCA) - the average per-class accuracy, which gives equal weight to both classes regardless of sample count - and the false positive rate (FPR) at fixed true positive rate (TPR) thresholds of 90\%, 80\%, and 70\%, following the protocol proposed in~\cite{Coskun2018n}. 
A high-quality classifier exhibits low, tightly clustered FPR values across all three thresholds.
For motion prediction, we report MPJPE in millimeters.

\subsection{Baseline Method Selection for Exercise Quality Classification}

Before applying MMD-NCA to the PROZIS classification task (Section~\ref{sec:results}.C), we summarize the baseline comparison presented by Coskun \textit{et al.}~\cite{Coskun2018n}, which motivates this choice of metric learning objective over temporal-alignment and pairwise/triplet-based alternatives.
Table~\ref{tab:fpr} presents FPR results on Human3.6M and CMU Mocap, comparing the MMD-NCA loss with established temporal alignment and metric learning baselines.
MMD-NCA achieves an FPR-90 of 32.66 on CMU and 38.42 on Human3.6M. Compared to DTW-based methods, the FPR reduction varies considerably by dataset and threshold, ranging from approximately 2.6 percentage points (vs.\ DCTW, Human3.6M FPR-80) to 18.8 percentage points (vs.\ CTW, CMU FPR-70), with larger and more consistent gains on CMU Mocap. Compared to triplet-based approaches, the reduction is more modest: roughly 6--8.5 percentage points on CMU Mocap, but only 0.8--4.4 percentage points on Human3.6M. The tight clustering of FPR values across the 90\%, 80\%, and 70\% thresholds further indicates a well-calibrated classifier.
Coskun et al.~\cite{Coskun2018n} further report that MMD-NCA achieves higher F1 score and Normalized Mutual Information than Triplet, Triplet+GOR, and N-Pair losses across embedding dimensions from 16 to 256 on both CMU and Human3.6M, reinforcing the FPR-based comparison above.

\begin{table}[t]
\caption{FPR (\%) at Fixed TPR Thresholds, as reported by Coskun \textit{et al.}~\cite{Coskun2018n} - Lower is Better}
\label{tab:fpr}
\centering
\small
\setlength{\tabcolsep}{3.5pt}
\begin{tabular}{lcccccc}
\toprule
\multirow{2}{*}{Method}
  & \multicolumn{3}{c}{CMU Mocap}
  & \multicolumn{3}{c}{Human3.6M} \\
\cmidrule(lr){2-4}\cmidrule(lr){5-7}
  & 90\% & 80\% & 70\%
  & 90\% & 80\% & 70\% \\
\midrule
DTW~\cite{Vintsyuk1968}             & 47.48 & 42.92 & 37.62 & 49.64 & 47.96 & 44.38 \\
MDDTW\cite{2name}         & 44.60 & 39.07 & 34.04 & 49.72 & 45.87 & 44.51 \\
CTW~\cite{3name}             & 46.02 & 40.96 & 39.11 & 47.63 & 43.10 & 42.18 \\

GDTW~\cite{Zhou2016}       
    & 45.61 & 39.95 & 35.24 & 46.06 & 42.72 & 40.04 \\
DCTW~\cite{Trigeorgis2018}       
    & 40.56 & 38.83 & 26.95 & 41.39 & 39.18 & 36.71 \\
    
Triplet~\cite{Schroff2015}     & 39.72 & 33.82 & 28.77 & 42.78 & 40.11 & 36.01 \\

Triplet+GOR~\cite{Zhang2017} 
    & 40.32 & 33.97 & 27.78 & 42.03 & 37.61 & 33.95 \\
    
N-Pair~\cite{Sohn2016}        & 40.11 & 32.35 & 26.16 & 40.46 & 39.56 & 36.52 \\
\textbf{MMD-NCA}~\cite{Coskun2018n}
  & \textbf{32.66} & \textbf{25.66} & \textbf{20.29}
  & \textbf{38.42} & \textbf{36.54} & \textbf{33.13} \\
\bottomrule
\end{tabular}
\end{table}

\subsection{Exercise Quality Classification on PROZIS Dataset}

\begin{figure}[t]
    \centering
    \begin{minipage}[b]{0.23\textwidth}
        \includegraphics[width=\textwidth]{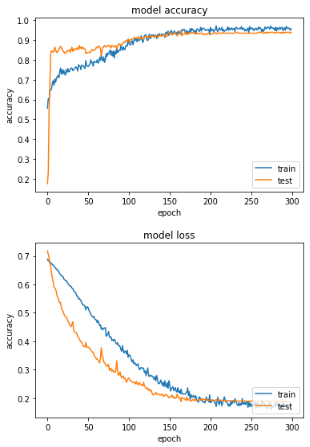}
        \centerline{(a) Squats}
        \label{fig:res0_squats}
    \end{minipage}
    \hfill
    \begin{minipage}[b]{0.23\textwidth}
        \includegraphics[width=\textwidth]{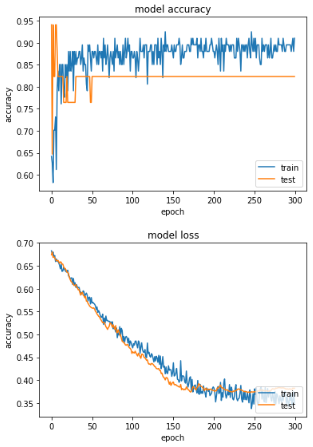}
        \centerline{(b) Sit-ups}
        \label{fig:res0_situps}
    \end{minipage}
    
    \vspace{0.15cm} 
    
    \begin{minipage}[b]{0.23\textwidth}
        \includegraphics[width=\textwidth]{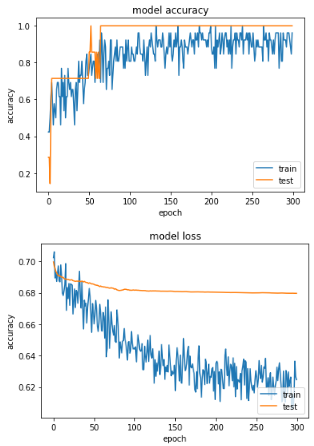}
        \centerline{(c) Push-ups}
        \label{fig:res0_pushups}
    \end{minipage}
    \hfill
    \begin{minipage}[b]{0.23\textwidth}
        \includegraphics[width=\textwidth]{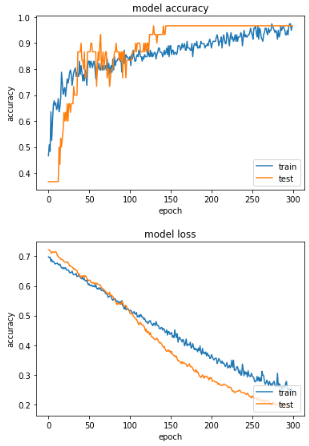}
        \centerline{(d) Jumping Jacks}
        \label{fig:res0_jumpjacks}
    \end{minipage}
    
    \caption{Training and validation accuracy and loss curves on the PROZIS dataset. The squat class (a) shows smooth convergence due to a high sample count, while classes with fewer samples, such as push-ups (c) and jumping jacks (d), have difficulty converging.}
    \label{fig:res0}
\end{figure}

For the squat class the model achieves
96.45\% mCA, with training and validation loss curves converging smoothly before epoch 200 and no signs of overfitting (see Fig.~\ref{fig:res0}). The self-attention mechanism focuses on the most discriminative phases of the repetition, particularly the lowest point of the squat descent, consistent with the attention behavior reported in \cite{Coskun2018n} for action retrieval.
For the remaining exercises (sit-ups, push-ups, and jumping jacks), the sample counts varied significantly, ranging from 70 to 300 sequences. Sit-ups showed mild underfitting and unstable training curves, while push-ups and jumping jacks failed to converge meaningfully, with validation accuracy collapsing to the class prior.

These failures are due to data scarcity and class imbalance rather than architectural limitations. The squat results show that the model is effective when sufficient data are available. Empirically, reliable convergence appears to require at least several hundred balanced sequences per exercise class. Inference per repetition takes approximately 20ms after extraction, which is compatible with per-repetition deployment once skeleton extraction is complete.

\subsection{Motion Prediction Backbone Selection}

Table~\ref{tab:mpjpe} summarizes the MPJPE benchmark as presented in~\cite{Xu2022n} for STARS and STS-GCN, together with LTD-50-25 and ConvSeq2Seq, presented in~\cite{Sofianos2021n}, to motivate this architectural choice. Our contribution is the application of this predictor within the joint-level feedback pipeline described in Section~\ref{sec:results}.E.

Table~\ref{tab:mpjpe} presents the MPJPE across all 15 Human3.6M actions. STARS consistently outperforms all baselines at every prediction horizon. For short-term prediction, STARS achieves 56.9 mm compared to 65.8 mm for STS-GCN, an improvement of 8.9 mm. For long-term prediction, STARS achieves 108.4 mm compared to 117.0 mm. Notably, STARS also outperforms LTD-50-25, which uses 50 context frames, while our method uses only 10 frames, demonstrating the advantage of the anchor-based generative formulation for long-horizon prediction.
As presented in \cite{Xu2022n}, STARS outperforms STS-GCN consistently across individual action categories, not only on average; the largest gains occur in high-variability, non-periodic actions such as Discussion and Posing, where the anchor-based generative formulation better captures multimodal upper-body motion, while periodic actions such as Walking show smaller but consistent improvements.

\begin{table}[t]

\caption{Average MPJPE (mm) on Human3.6M~\cite{Sofianos2021n, Xu2022n} - Lower is Better}

\label{tab:mpjpe}
\centering
\small
\setlength{\tabcolsep}{2.5pt}
\begin{tabular}{lcccccccc}
\toprule
\multirow{2}{*}{Method}
  & \multicolumn{4}{c}{Short-term (ms)}
  & \multicolumn{4}{c}{Long-term (ms)} \\
\cmidrule(lr){2-5}\cmidrule(lr){6-9}
  & 80 & 160 & 320 & 400
  & 560 & 720 & 880 & 1000 \\
\midrule
ConvSeq2Seq 
  & 16.6 & 33.3 & 53.9 & 61.2
  & 90.7 & 104.7 & 116.7 & 124.2 \\
LTD-50-25 
  & 11.2 & 23.4 & 47.9 & 58.9
  & 79.6 &  93.6 & 105.2 & 112.4 \\
STS-GCN 
  & 13.5 & 27.7 & 54.4 & 65.8
  & 85.0 &  98.3 & 108.9 & 117.0 \\
\textbf{STARS} 
  & \textbf{10.0} & \textbf{21.8} & \textbf{45.7} & \textbf{56.9}
  & \textbf{75.8} & \textbf{89.3} & \textbf{100.8} & \textbf{108.4} \\
\bottomrule
\end{tabular}
\end{table}

The choice of input parameterization significantly affects prediction accuracy: using raw 3D Cartesian coordinates instead of the exponential map representation reduces short-term accuracy by approximately 23\% and long-term accuracy by approximately 34\%~\cite{Sofianos2021n}. The exponential map avoids gimbal locking but requires inverse and forward kinematics functions, which add substantial preprocessing complexity. For the rehabilitation application, where prediction horizons are intentionally short and feedback resolution is at the joint rather than the sub-centimeter level, the Cartesian representation is a pragmatically acceptable trade-off.

\subsection{Qualitative Feedback Analysis}

\begin{figure*}[!htbp] 
    \centering
    \includegraphics[width=0.19\textwidth]{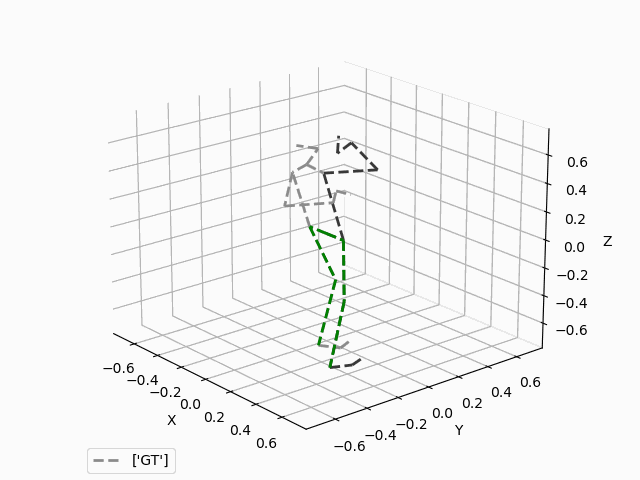}\hfill
    \includegraphics[width=0.19\textwidth]{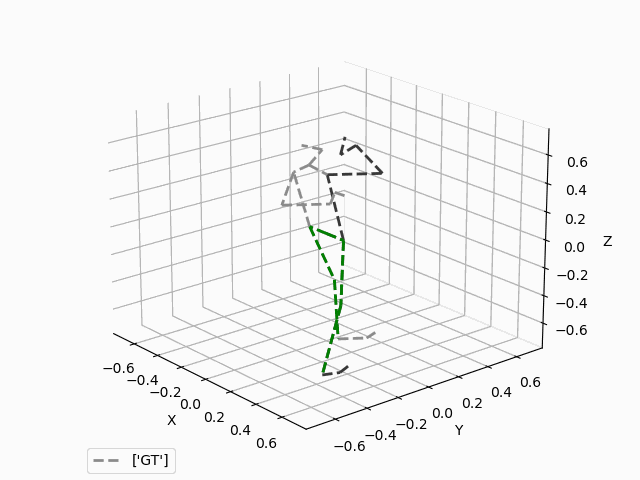}\hfill
    \includegraphics[width=0.19\textwidth]{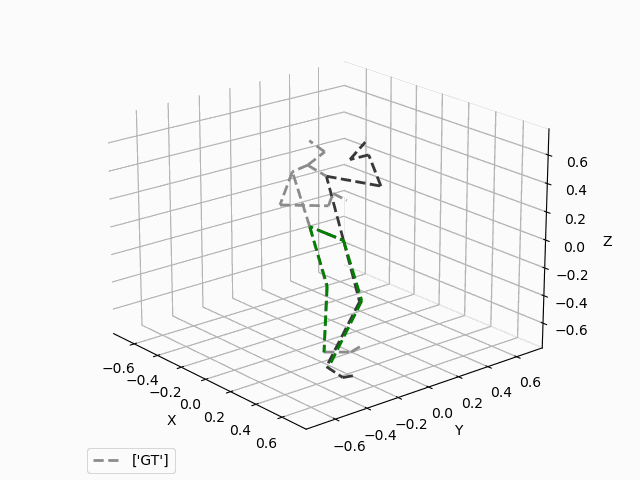}\hfill
    \includegraphics[width=0.19\textwidth]{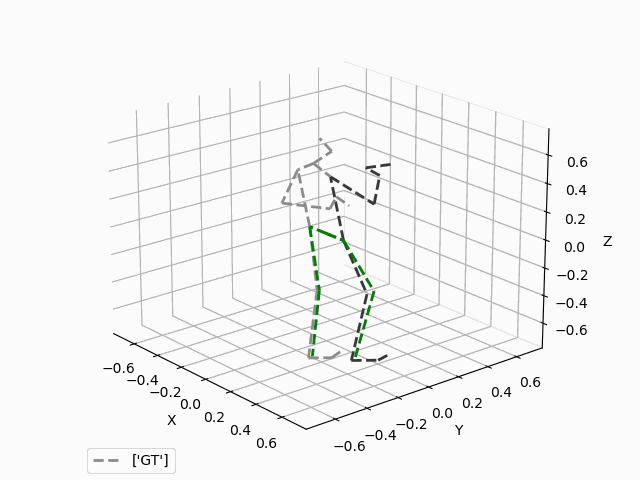}\hfill
    \includegraphics[width=0.19\textwidth]{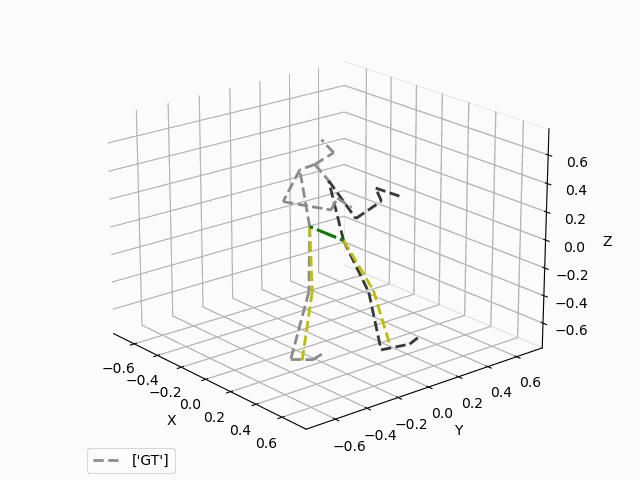}
    
    \caption{Qualitative visual feedback generated by STS-GCN for a walking sequence. Five evenly spaced frames sampled from a continuous 25-frame ($\approx$1s)  prediction horizon are shown. The predicted skeleton is color-coded by per-joint position error (green: low; yellow: moderate; red: high) against the ground truth (dotted line). Accuracy remains high initially before open-loop error naturally accumulates in the later frames.}
    \label{fig:walking_sequence}
\end{figure*}

The per-joint error signal was qualitatively evaluated on a walking sequence from Human3.6M, using STS-GCN with 3D Cartesian input and a 25-frame prediction horizon (approximately 1 second), as shown in Fig.~\ref{fig:walking_sequence}. During correctly performed linear walking, leg joints remained mostly in the green and yellow error bands for the first 15 frames, with error accumulating in the final frames as expected due to the open-loop nature of the prediction. This demonstrates that the feedback mechanism correctly highlights deviations only when they become genuinely significant, rather than producing noisy
false alerts throughout the sequence.

\begin{figure*}[!ht]
    \centering
    \includegraphics[width=0.19\textwidth]{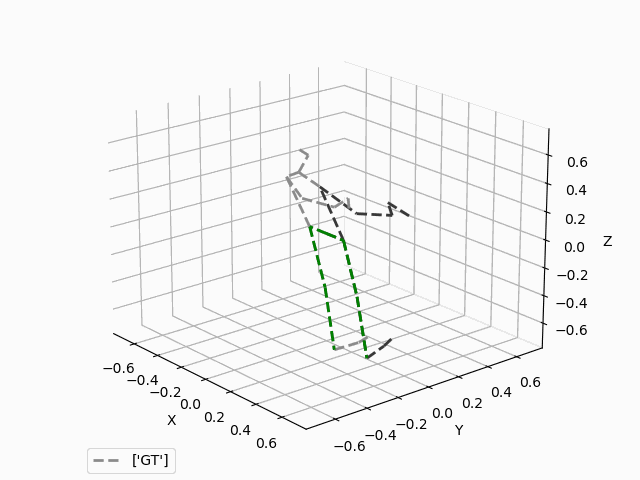}\hfill
    \includegraphics[width=0.19\textwidth]{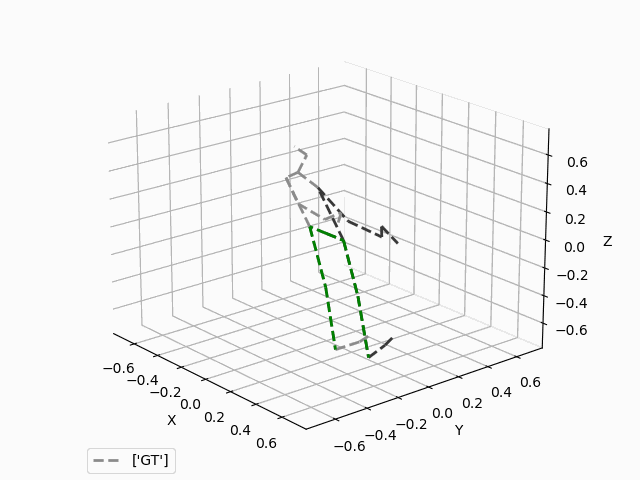}\hfill
    \includegraphics[width=0.19\textwidth]{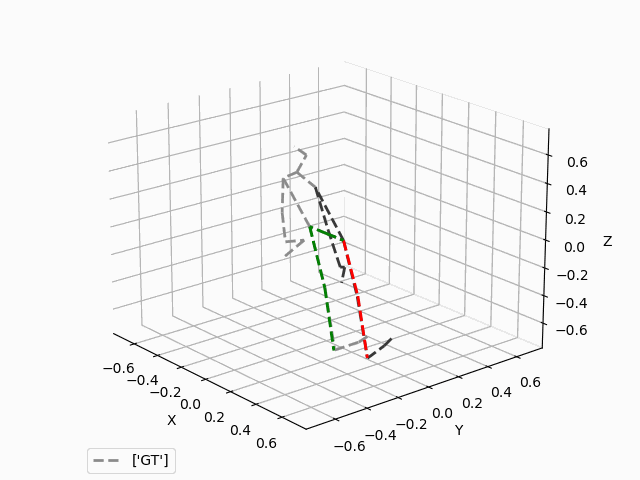}\hfill
    \includegraphics[width=0.19\textwidth]{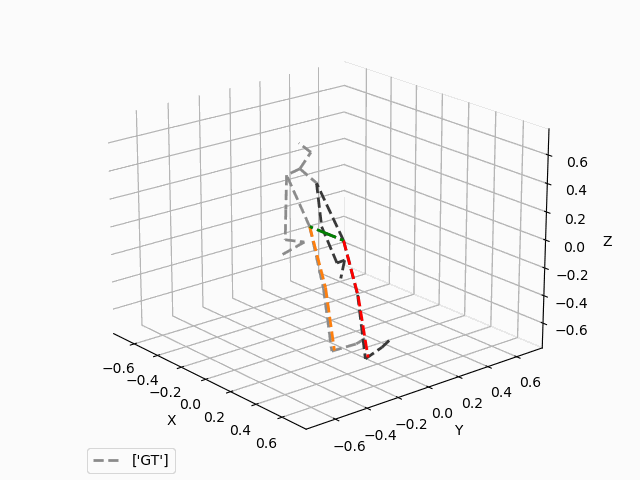}\hfill
    \includegraphics[width=0.19\textwidth]{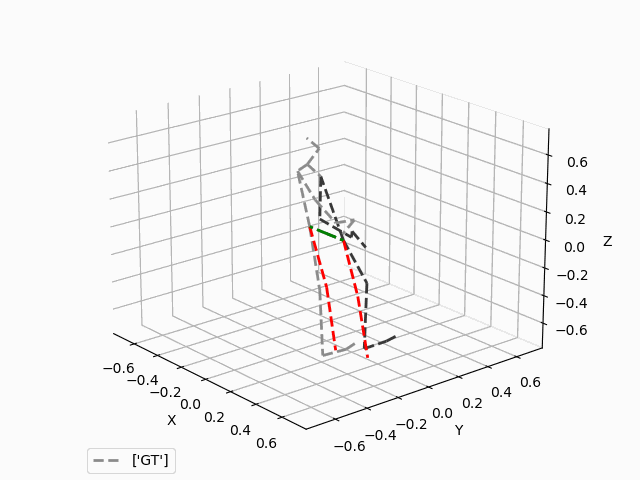}

    \caption{Effect of context quality on joint-level feedback. Five evenly 
    spaced frames from a 25-frame ($\approx$1s) prediction horizon are shown for a walking sequence that begins from a stationary pose. High initial errors (orange/red joints) decrease rapidly once active motion context accumulates, motivating a minimum buffer of 10 frames (400\,ms) of active movement before activating the feedback overlay.}
    \label{fig:context_quality}
\end{figure*}

Robustness to execution speed was also evaluated by subsampling the same walking sequence to 50\% and 25\% of its original frame rate, which increases the spatial distance between joints across consecutive frames. The predictor adapted its output to the altered temporal scale without retraining or explicit dynamic time warping, consistent with the tempo-invariance design goal described in Section~\ref{sec:pipeline}.

Context quality is essential: sequences that begin with a static pose result in significantly higher joint errors in the initial frames, as shown in Fig.~\ref{fig:context_quality}. This occurs because the predictor lacks an active motion signal to condition on and defaults to a near-stationary prediction, which diverges immediately once the subject starts moving. Error rapidly decreases once sufficient active-movement context accumulates. This motivates buffering at least 10 frames (400\,ms) of active movement before activating the feedback overlay, and suggests that triggering the feedback only after the first full repetition begins is a practical deployment strategy.

End-to-end inference time from preprocessed skeleton input to rendered 3D overlay is 2--5\,s in Python on an RTX~2060, making the system suitable for per-repetition rather than per-frame feedback. Optimization via C++ or ONNX deployment is left for future work.

\section{Conclusion}
\label{sec:conclusion}

We presented a two-module deep learning pipeline that combines exercise quality assessment with joint-level deviation signaling to support telerehabilitation. The pipeline uses a self-attentive BiLSTM classifier using
MMD-NCA metric learning for holistic exercise quality classification, and a graph-based short-term motion predictor for spatially localized feedback. Evaluated on established benchmarks, the classifier achieves 96.45\% mean-class accuracy on squat sequences from the PROZIS dataset,
while the STARS predictor achieves 75.8\,mm MPJPE at 560\,ms on Human3.6M, outperforming graph and recurrent baselines across all prediction horizons. Both modules operate on marker-free RGB video without specialized hardware. The primary contribution is the integration of these two complementary signals – holistic quality labels and per-joint deviation maps – into system design
intended for deployment on assistive and social robots.
 

The current work has four limitations that scope the present contribution honestly and define directions for future research.

First, the two modules are trained and evaluated on separate datasets, and no end-to-end evaluation of the complete recognition--prediction--feedback loop has been conducted. As noted in Section~\ref{sec:data}, this is in part a current data constraint: PROZIS lacks the kinematic-tree information required by the graph-based predictor, and Human3.6M contains no rehabilitation exercises or quality annotations.
Second, the joint-level error signal (Section~\ref{sec:mp}.D) is derived from motion-prediction error rather than therapist-annotated deviation ground truth. We present the JPE signal as a candidate feedback mechanism, with clinical validation identified as essential future work.
Third, end-to-end inference takes 2--5\,seconds, restricting the system to per-repetition rather than frame-level or continuous feedback.

This study establishes the feasibility of the system by validating its individual components—a necessary first step before testing the complete setup with users, therapists, and robotic hardware. Building on this foundation, future work will focus on end-to-end validation for rehabilitation exercises. We plan to improve anatomical accuracy by integrating kinematic constraints, validate our joint-level feedback against physiotherapist-annotated data, and prospective evaluation with users and clinicians in a human-robot interaction setting.

\section*{Acknowledgments}

This work was funded by national funds through FCT - Foundation for Science and Technology, I.P., under the grant UID/00048/2025 (DOI: 10.54499/UIDB/00048/2025).

\balance
\bibliographystyle{IEEEtran}
\bibliography{ref.bib}

@article{Rigamonti2020,
   author = {Lia Rigamonti and Urs Vito Albrecht and Christoph Lutter and Mathias Tempel and Bernd Wolfarth and David Alexander Back and David Alexander Back},
   doi = {10.1249/jsr.0000000000000704},
   issn = {15378918},
   issue = {4},
   journal = {Current Sports Medicine Reports},
   month = {4},
   pages = {157-163},
   pmid = {32282462},
   publisher = {Lippincott Williams and Wilkins},
   title = {Potentials of digitalization in sports medicine: A narrative review},
   volume = {19},
   year = {2020}
}

@article{Liao2020,
   author = {Yalin Liao and Aleksandar Vakanski and Min Xian},
   doi = {10.1109/TNSRE.2020.2966249},
   issn = {15580210},
   issue = {2},
   journal = {IEEE Transactions on Neural Systems and Rehabilitation Engineering},
   month = {2},
   pages = {468-477},
   pmid = {31940544},
   publisher = {Institute of Electrical and Electronics Engineers Inc.},
   title = {A Deep Learning Framework for Assessing Physical Rehabilitation Exercises},
   volume = {28},
   year = {2020}
}

@article{Coskun2018n,
   author = {Huseyin Coskun and David Joseph Tan and Sailesh Conjeti and Nassir Navab and Federico Tombari},
   journal = {Lecture Notes in Computer Science (including subseries Lecture Notes in Artificial Intelligence and Lecture Notes in Bioinformatics)},
   month = {8},
   pages = {693-710},
   publisher = {Springer Verlag},
   title = {Human Motion Analysis with Deep Metric Learning},
   volume = {11218 LNCS},
   url = {http://arxiv.org/abs/1807.11176},
   year = {2018}
}

@article{Lin2017,
   author = {Zhouhan Lin and Minwei Feng and Cicero Nogueira dos Santos and Mo Yu and Bing Xiang and Bowen Zhou and Yoshua Bengio},
   journal = {5th International Conference on Learning Representations, ICLR 2017 - Conference Track Proceedings},
   month = {3},
   publisher = {International Conference on Learning Representations, ICLR},
   title = {A Structured Self-attentive Sentence Embedding},
   url = {http://arxiv.org/abs/1703.03130},
   year = {2017}
}

@article{Sofianos2021n,
   author = {Theodoros Sofianos and Alessio Sampieri and Luca Franco and Fabio Galasso},
   journal = {Proceedings of the IEEE International Conference on Computer Vision},
   month = {10},
   pages = {11189-11198},
   publisher = {Institute of Electrical and Electronics Engineers Inc.},
   title = {Space-Time-Separable Graph Convolutional Network for Pose Forecasting},
   url = {http://arxiv.org/abs/2110.04573},
   year = {2021}
}

@article{Xu2022n,
   author = {Sirui Xu and Yu Xiong Wang and Liang Yan Gui},
   doi = {10.1007/978-3-031-20047-2_15},
   isbn = {9783031200465},
   issn = {16113349},
   journal = {Lecture Notes in Computer Science (including subseries Lecture Notes in Artificial Intelligence and Lecture Notes in Bioinformatics)},
   pages = {251-269},
   publisher = {Springer, Cham},
   title = {Diverse Human Motion Prediction Guided by Multi-level Spatial-Temporal Anchors},
   year = {2022}
}

@article{Ionescu2014n,
   author = {Catalin Ionescu and Dragos Papava and Vlad Olaru and Cristian Sminchisescu},
   doi = {10.1109/TPAMI.2013.248},
   issn = {01628828},
   issue = {7},
   journal = {IEEE Transactions on Pattern Analysis and Machine Intelligence},
   pages = {1325-1339},
   pmid = {26353306},
   publisher = {IEEE Computer Society},
   title = {Human3.6M: Large scale datasets and predictive methods for 3D human sensing in natural environments},
   volume = {36},
   year = {2014}
}

@article{Jain2016n,
   author = {Ashesh Jain and Amir R. Zamir and Silvio Savarese and Ashutosh Saxena},
   journal = {Proceedings of the IEEE Computer Society Conference on Computer Vision and Pattern Recognition},
   month = {4},
   pages = {5308-5317},
   publisher = {IEEE Computer Society},
   title = {Structural-RNN: Deep Learning on Spatio-Temporal Graphs},
   volume = {2016-December},
   url = {http://arxiv.org/abs/1511.05298},
   year = {2016}
}

@misc{batista2019prozis,
  author       = {Batista, J.},
  title        = {Prozis Challenge},
  year         = {2019},
  howpublished = {\url{https://www.isr.uc.pt/index.php/projects/past-projects?task=showprojects.show%28%29&idProject=219}},
  note         = {Online; accessed 2025}
}

@article{Ferreira2022,
   author = {Bruno Ferreira and Paulo Menezes and Jorge Batista},
   doi = {10.1109/ICIP46576.2022.9897194},
   isbn = {9781665496209},
   issn = {15224880},
   journal = {Proceedings - International Conference on Image Processing, ICIP},
   pages = {3470-3474},
   publisher = {IEEE Computer Society},
   title = {TRANSFORMERS FOR WORKOUT VIDEO SEGMENTATION},
   url = {https://ieeexplore.ieee.org/document/9897194},
   year = {2022}
}

@article{Sutherland2021n,
   author = {Danica J. Sutherland and Hsiao-Yu Tung and Heiko Strathmann and Soumyajit De and Aaditya Ramdas and Alex Smola and Arthur Gretton},
   journal = {5th International Conference on Learning Representations, ICLR 2017 - Conference Track Proceedings},
   month = {1},
   publisher = {International Conference on Learning Representations, ICLR},
   title = {Generative Models and Model Criticism via Optimized Maximum Mean Discrepancy},
   url = {http://arxiv.org/abs/1611.04488},
   year = {2021}
}

@misc{CMUd,
   title = {Carnegie Mellon University - CMU Graphics Lab - motion capture library},
   url = {https://mocap.cs.cmu.edu/}
}

@article{Mao2020,
   author = {Wei Mao and Miaomiao Liu and Mathieu Salzmann and Hongdong Li},
   journal = {Proceedings of the IEEE International Conference on Computer Vision},
   month = {7},
   pages = {9488-9496},
   publisher = {Institute of Electrical and Electronics Engineers Inc.},
   title = {Learning Trajectory Dependencies for Human Motion Prediction},
   url = {http://arxiv.org/abs/1908.05436},
   year = {2020}
}

@article{vintsyuk1968speech,
  author  = {Vintsyuk, T. K.},
  title   = {Speech Discrimination by Dynamic Programming},
  journal = {Cybernetics},
  volume  = {4},
  number  = {1},
  pages   = {52--57},
  year    = {1968},
  note    = {Russian original: Kibernetika 4(1):81--88 (1968)}
}

@article{Keogh2001,
   author = {Eamonn J. Keogh and Michael J. Pazzani},
   doi = {10.1137/1.9781611972719.1},
   journal = {Proceedings},
   month = {4},
   pages = {1-11},
   publisher = {Society for Industrial \& Applied Mathematics (SIAM)},
   title = {Derivative Dynamic Time Warping},
   url = {/doi/pdf/10.1137/1.9781611972719.1?download=true},
   year = {2001}
}

@article{Hadsell2006n,
   author = {Raia Hadsell and Sumit Chopra and Yann LeCun},
   doi = {10.1109/CVPR.2006.100},
   isbn = {0769525970},
   issn = {10636919},
   journal = {Proceedings of the IEEE Computer Society Conference on Computer Vision and Pattern Recognition},
   pages = {1735-1742},
   title = {Dimensionality reduction by learning an invariant mapping},
   volume = {2},
   url = {https://ieeexplore.ieee.org/document/1640964},
   year = {2006}
}

@article{Schroff2015n,
   author = {Florian Schroff and Dmitry Kalenichenko and James Philbin},
   doi = {10.1109/CVPR.2015.7298682},
   isbn = {9781467369640},
   issn = {10636919},
   journal = {Proceedings of the IEEE Computer Society Conference on Computer Vision and Pattern Recognition},
   month = {10},
   pages = {815-823},
   publisher = {IEEE Computer Society},
   title = {FaceNet: A unified embedding for face recognition and clustering},
   volume = {07-12-June-2015},
   url = {https://ieeexplore.ieee.org/document/7298682},
   year = {2015}
}

@article{Martinez2017n,
   author = {Julieta Martinez and Michael J. Black and Javier Romero},
   journal = {Proceedings - 30th IEEE Conference on Computer Vision and Pattern Recognition, CVPR 2017},
   month = {5},
   pages = {4674-4683},
   publisher = {Institute of Electrical and Electronics Engineers Inc.},
   title = {On human motion prediction using recurrent neural networks},
   volume = {2017-January},
   url = {http://arxiv.org/abs/1705.02445},
   year = {2017}
}

@article{Yan2018n,
   author = {Sijie Yan and Yuanjun Xiong and Dahua Lin},
   journal = {32nd AAAI Conference on Artificial Intelligence, AAAI 2018},
   month = {1},
   pages = {7444-7452},
   publisher = {AAAI press},
   title = {Spatial Temporal Graph Convolutional Networks for Skeleton-Based Action Recognition},
   url = {http://arxiv.org/abs/1801.07455},
   year = {2018}
}

@article{Vintsyuk1968,
   author = {T. K. Vintsyuk},
   doi = {10.1007/BF01074755/METRICS},
   issn = {15738337},
   issue = {1},
   journal = {Cybernetics},
   month = {1},
   pages = {52-57},
   publisher = {Kluwer Academic Publishers-Plenum Publishers},
   title = {Speech discrimination by dynamic programming},
   volume = {4},
   url = {https://link.springer.com/article/10.1007/BF01074755},
   year = {1968}
}

@article{2name,
   author = {Jiangyuan Mei and Meizhu Liu and Yuan-Fang Wang and Huijun Gao},
   title = {Learning a Mahalanobis Distance based Dynamic Time Warping Measure for Multivariate Time Series Classification}
}

@article{3name,
   author = {Feng Zhou and Fernando De La Torre},
   title = {Canonical Time Warping for Alignment of Human Behavior},
}

@article{Zhou2016,
  author  = {Zhou, Feng and De la Torre, Fernando},
  title   = {Generalized Canonical Time Warping},
  journal = {IEEE Transactions on Pattern Analysis and Machine Intelligence},
  volume  = {38}, number = {2}, pages = {279--294}, year = {2016}
}

@article{Trigeorgis2018,
   author = {George Trigeorgis and Mihalis A. Nicolaou and Bjorn W. Schuller and Stefanos Zafeiriou},
   doi = {10.1109/TPAMI.2017.2710047},
   issn = {0162-8828},
   issue = {05},
   journal = {IEEE Transactions on Pattern Analysis and Machine Intelligence},
   month = {5},
   pages = {1128-1138},
   pmid = {28613160},
   publisher = {IEEE Computer Society},
   title = {Deep Canonical Time Warping for Simultaneous Alignment and Representation Learning of Sequences},
   volume = {40},
   year = {2018}
}

@article{Schroff2015,
   author = {Florian Schroff and Dmitry Kalenichenko and James Philbin},
   doi = {10.1109/CVPR.2015.7298682},
   isbn = {9781467369640},
   issn = {10636919},
   journal = {Proceedings of the IEEE Computer Society Conference on Computer Vision and Pattern Recognition},
   month = {10},
   pages = {815-823},
   publisher = {IEEE Computer Society},
   title = {FaceNet: A unified embedding for face recognition and clustering},
   volume = {07-12-June-2015},
   url = {https://ieeexplore.ieee.org/document/7298682},
   year = {2015}
}

@article{Zhang2017,
   author = {Xu Zhang and Felix X. Yu and Sanjiv Kumar and Shih Fu Chang},
   doi = {10.1109/ICCV.2017.492},
   isbn = {9781538610329},
   issn = {15505499},
   journal = {IEEE International Conference on Computer Vision},
   month = {12},
   pages = {4605-4613},
   publisher = {Institute of Electrical and Electronics Engineers Inc.},
   title = {Learning Spread-Out Local Feature Descriptors},
   volume = {2017-October},
   year = {2017}
}

@article{Sohn2016,
   author = {Kihyuk Sohn},
   journal = {Neural Information Processing Systems},
   title = {Improved Deep Metric Learning with Multi-class N-pair Loss Objective},
   year = {2016}
}

\vfill

\end{document}